\documentclass[10pt,twocolumn,letterpaper]{article}

\usepackage{cvpr}      % To produce the REVIEW version
\usepackage{graphicx}
\usepackage{amsmath}
\usepackage{amssymb}
\usepackage{booktabs}
\usepackage{dsfont}
\usepackage[pagebackref,breaklinks,colorlinks]{hyperref}

\usepackage[capitalize]{cleveref}
\crefname{section}{Sec.}{Secs.}
\Crefname{section}{Section}{Sections}
\Crefname{table}{Table}{Tables}
\crefname{table}{Tab.}{Tabs.}

\def\cvprPaperID{6672} % *** Enter the CVPR Paper ID here
\def\confName{CVPR}
\def\confYear{2022}

\begin{document}

% \title{\LaTeX\ Author Guidelines for \confName~Proceedings}
\title{StyleM: Stylized Metrics for Image Captioning Built with Contrastive $N$-grams}
\author{Chengxi Li\\
University of Kentucky\\
329 Rose St, Lexington, KY 40508\\
\and
Brent Harrison\\
University of Kentucky\\
329 Rose St, Lexington, KY 40508\\
}
\maketitle

%%%%%%%%% ABSTRACT
\begin{abstract}
In this paper, we build two automatic evaluation metrics for evaluating the association between a machine-generated caption and a ground truth stylized caption: OnlyStyle and StyleCIDEr  \footnote{code will be available at https://github.com/***}.
%for evaluating the association between a caption and a given style. 
OnlyStyle can be used to score any caption versus a given style. 
%generated caption or ground truth caption with a given style. 
StyleCIDEr supports scoring the similarity of two compared captions with respect to their styles. We evaluate these two metrics using three stylized captioning methods trained on the PERSONALITY-CAPTIONS and FlickrStyle10K datasets: UPDOWN, MULTI-UPDOWN, and SVinVL. %And we re-evaluate three popular stylized image captioning under these metrics: UPDOWN, MULTI-UPDOWN and SVinVL. 
We also perform a human study to explore how well each caption aligns with human judgments in similar situations. 
\end{abstract}

%%%%%%%%% BODY TEXT
\section{Introduction}
While classic image captioning approaches show deep understanding of image composition and language construction, they often lack elements that make communication distinctly human. 
To address this issue, some researchers have tried to add personality to image captioning in order to generate stylized captions. 
With this new type of task, researchers \cite{gan2017stylenet,zhao2020memcap,guo2019mscap,shuster2019engaging, li20213m} often use automatic NLP evaluation metrics, like BLEU \cite{papineni2002bleu}, ROUGE-L\cite{lin-2004-rouge}, CIDEr\cite{lin-2004-rouge}, and SPICE\cite{anderson2016spice} for evaluating the accuracy of the generations and a trained classifier for evaluating the style components' contributions.
However, a trained classifier for evaluation is constrained by its own accuracy once the number of classes scales up. 
Changing datasets will also require another benchmark model to be trained and agreed upon by different researchers. 
If individual researchers trained their own classifiers for evaluation, then it is difficult to achieve a fair comparison between techniques as differences in training protocols could lead to differences in performance that are difficult to reproduce. 
\begin{figure}
\centering
 \includegraphics[width= \columnwidth ,height=7cm]{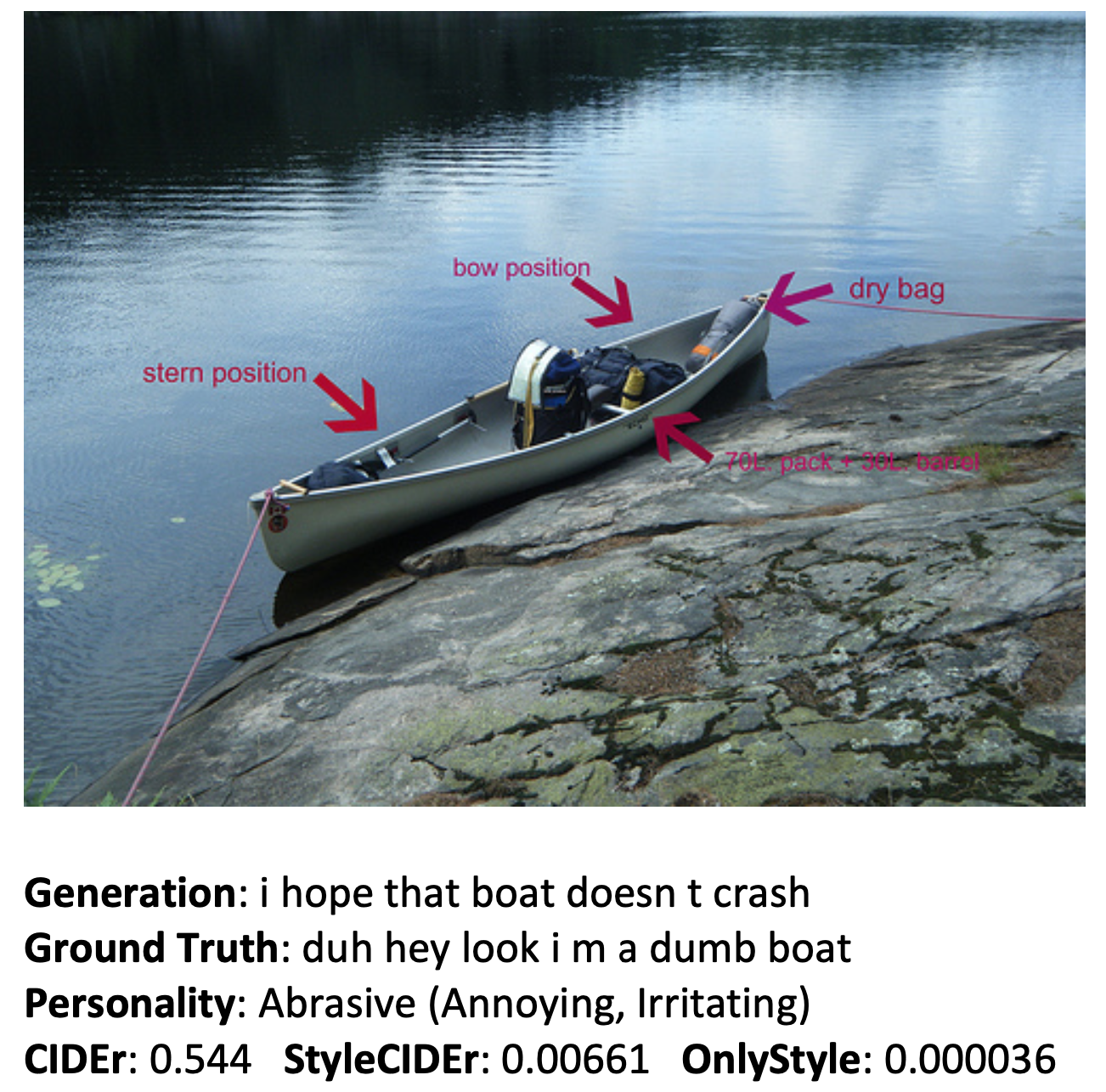}
\caption{An image generated caption for the personality "Abrasive (Annoying, Irritating)". The generated caption are measured using CIDEr, StyleCIDEr, OnlyStyle scores. CIDEr fails to reflect styles}
\label{fig:ciderexample}
\end{figure}

Automatic evaluation metrics such as BLEU, CIDEr are based on measuring the alignment between generated captions and reference captions using $n$-grams.
These metrics are frequently used to evaluate language generation models because they don't require training, which makes them suitable for providing stable evaluations across different models. 

Given that stylized captions must describe the image in question as well as produce a sentence that aligns with a given style, it may be difficult for these types of metrics to give an accurate measure of the quality of a stylized captioning system. 
If the CIDEr score, for example, is high for a stylized image caption, a possible explanation is because it can accurately recreate elements of a caption that describe the image. In these instances, it may fail to encompass the desired stylized elements of the caption. 
This can be seen happening in Figure \ref{fig:ciderexample}, where the CIDEr score is high due to the 1-grams ``i" and ``boat" aligning with ground truth. However, these two words could hardly reflect the desired personality. Thus, the CIDEr-like scores may result in accurate evaluations for stylized captioning models.

For a metric to accurately evaluate the quality of a stylized image caption, it must be able to reason about how the words in said caption contribute to its overall style. To do that, we propose to build automatic evaluation metrics from stylized captioning datasets and use them to measure how words contribute to the overall style of automatically generated captions. 

To do this, we must first address the following questions: 1. Do we need a reference caption to measure a generated caption's ability to capture style? 2. Does the metric align with human perceptions of a sentence's style?

To address the first question, we recognize that there are benefits to creating a metric that requires a reference caption. 
The presence of a reference caption would allow us to directly compare if the generated stylized caption is similar to a known ground truth. 
There are situations, however, where a ground truth is not easily obtainable. 
In these situations, one would prefer a metric that did not require a reference sentence. 
Considering this, we will design two metrics that are both based on the $n$-gram methods described earlier. 
The first will evaluate how well a generated caption captures a given style without the need for a reference sentence. We will refer to this metric as \textbf{OnlyStyle}. 
The second metric will be based on the popular CIDEr metric, but we will bias its measures to place more emphasis on common stylized words to better measure the quality of a stylized caption.
We will refer to this metric as \textbf{StyleCIDEr}.

For the second question, we do want our metrics align human's judgment. CIDEr \cite{vedantam2015cider} and Self-CIDEr \cite{wang2019describing} have been demonstrated to match human preferences when comparing two captions. However, evaluating styles can be difficult as people may have different thoughts on what constitutes a particular style. For example, the data used to train a stylized image captioning model may indicate that words such as ``I love'' indicates a ``Happy'' style while a human judge may disagree. 

Thus, in this paper we choose to score stylized traits based on the dataset used to train the captioning model. 
While this is our primary evaluation, we do also perform a human subjects evaluation to see if our metric aligns with human preferences. 

\section{Related Work}

\textbf{Stylized Image Captioning Models}
Several researchers have built stylized captioners \cite{gan2017stylenet,chen2018factual,guo2019mscap,zhao2020memcap} using the FlickrStyle10K dataset which includes Humorous and Romantic personalities. Some of these works use unpaired datasets and LSTM\cite{greff2016lstm} variants for caption models.
Shuster \textit{et al.} \cite{shuster2019engaging} released the PERSONALITY-CAPTIONS dataset containing 215 personalities in 2019. They improved on past approaches for stylized captioning by improving input features and exploring various state-of-the-art image captioning models~\cite{xu2015show,anderson2018bottom}. Among them, the best performing generative caption model is UPDOWN and the best features for representing image input are generated using ResNext \cite{xie2017aggregated}.
Li \textit{et al.} extended the UPDOWN model into the Multi-UPDOWN model which incorporates both visual and text information as inputs. Oscar~\cite{li2020oscar} also uses fusion over text and visual features in conjunction with a transformer model trained using contrastive loss. 
VinVL\cite{zhang2021vinvl} outperforms Oscar by utilizing better text features and has since become the state-of-art for image captioning work. 

In this paper, we will evaluate some of these frameworks using our proposed metrics. 
Specifically, we evaluate UPDOWN, Multi-UPDOWN, and a VinVL model fine-turned on a stylized captioning dataset (we call this model SVinVL for convenience). 

This will enable us to demonstrate how our proposed metric can be used to evaluate stylized captioning models. \\
\textbf{Common Metrics Used in stylized Image Captioning}
In general, there are two types of metrics used to evaluate stylized image captioning models: trained models and automatic NLP evaluation metrics. 

There are many types of trained models that can be used for stylized caption evaluation. These include text classifiers trained to identify the style of a sentence, or a model like BertScore~\cite{zhang2019bertscore}, which uses a trained language model to compare generated text against a reference sentence. 
One of the main limitations of using a trained model as an evaluation metric is that it may require retraining to use on a different dataset. 
Thus, metrics based on trained models may require significant effort if used by different researchers on different datasets or in different domains. 
We would like to develop a general metric that performs consistently once the dataset is fixed. 
Thus, we feel that our approach is more similar to automatic NLP evaluation metrics. 
Many automated NLP evaluation metrics rely on $n$-gram matching precision or recall, and are not necessarily designed with image captioning in mind. 
Examples of these metrics include BLEU\cite{papineni2002bleu}, METEOR \cite{banerjee-lavie-2005-meteor} and ROUGE-L \cite{lin-2004-rouge}.
SPICE \cite{anderson2016spice} and CIDEr\cite{vedantam2015cider} are designed specifically for evaluating image captioning. 
SPICE depends on scene graph matching, which is heavily influenced by the accuracy of the parsing results. 
CIDEr averages the cosine similarities between two caption vectors across 1-4 grams, where vectors are calculated using TF-IDF. 
Since CIDEr is generally considered to be effective at sensing semantic changes and representing a human's preference \cite{wang2019describing}, Wang \textit{et al.} created Self-CIDEr \cite{wang2019describing} to measure the diversity of caption generations by using CIDEr as a kernel function. 
While effective on general captioning tasks, CIDEr can struggle to identify the stylized elements of a caption. \\
\textbf{N-Gram Weighting Schemes} Since our metrics are based on $n$-gram methods, we will discuss several popular term weighting schemes\cite{joachims1996probabilistic,martineau2009delta}. Term Frequency-Inverse Document Frequency (TF-IDF) is effective at scoring salient $n$-grams; however, due to the fact they punish stop words, they could weight words such as ``not" as 0 with the help of inverse document frequency if ``not" appears in every style. To us, this is limiting since this ignores the term frequency of ``not" in different stylized corpora and weights it the same in a "Happy" style and an "Angry" style. Delta TF-IDF\cite{martineau2009delta} uses the difference of TF-IDF scores between a positive dataset and negative dataset. 
This may help to extract more stylized words than TF-IDF, but does not fix any of the underlying issues with TF-IDF. We will address this by using the average difference of term frequency probabilities between a positive corpus and negative corpus.

\section{Method}
In this paper we introduce two metrics for automatically evaluating stylized image captions. 
To accomplish this,we divide the task into three parts. 
First, for each style in a dataset we calculate a contrastive $n$-gram (\textit{CNG}) score for each $n$-gram in the corpus, which gives a high score to $n$-gram representative of the given style and a low score for those that are not. 

We use this $n$-gram score to represent the associations between a given style and an $n$-gram. Secondly, we use these calculated $n$-gram scores to build our first metric, which takes a caption and a style as input and outputs the a score measuring how much this caption exhibits this style. We refer to this metric as \textit{OnlyStyle}. Thirdly, we construct a second metric, \textit{StyleCIDEr}, that measures whether the stylized elements of a sentence align with a reference caption or not. 
This metric uses CIDEr as a kernel function, but shifts the $n$-gram vector space. 
We will talk about each of these three concepts in more detail below. 
\subsection{Contrastive N-gram Score}
We build our metric based on $n$-gram comparisons, which means we will first calculate a score for each $n$-gram in a caption that measures how aligned it is with a given style. Generally, we would like score an $n$-gram high if it only appears in the dataset associated with current style and if it appears frequently in this dataset. That is to say, if an $n$-gram has high contrastive $n$-gram score on style $p$, we would think this $n$-gram contains good characteristics for this style. Suppose we have a $n$-gram $t$, $ n\in{\left \{ 1,2,3,4 \right \}}$, and we would like to calculate how close this $t$ with a given style $p$.
We calculate the score $s_{p,n}$ for this $n$-gram $t$ under style $p$  as follows:
\begin{align}
&f_{p,n} = | \left \{ d\in D_{p}: t\in d \right \}| \label{eq:ngram1}\\
& ECDF(f_{p,n}) =\frac{1}{\|F_{p,n}\|}\sum_{x_{i}\in F_{p,n}} \mathds{1}_{x_{i}<=f_{p,n}} \label{eq:ngram2}\\
& occur\_num = | \left \{ s\in S: t\in d, d \in D_{s} \right \}| \label{eq:ngram3} \\
& CNG_{p,n} = \frac{1}{\|S\|}\sum_{q\in S, q!=p}\frac{ECDF(f_{p,n})-ECDF(f_{q,i})}{occur\_num}\label{eq:ngram4}
\end{align}
Equation \ref{eq:ngram1} calculates the document frequency for $t$ in $D_{p}$ where $t$ is a $n$-gram term and $D_{p}$ is the corpus that includes all sentences related to the style $p$. We intend to compare the $t$'s representations across different styles to see whether $t$ is salient in current style $p$. Since document frequency like $f_{p,n}$ is heavily influenced by the number of documents in $D_{p}$, %for example, 1000 documents could have 5 documents containing $t$ while 10000 documents could have 20 documents containing $t$, 
we need to convert $f_{p,n}$ into probability.
%\brent{Think about how to divide this sentence into maybe two (have the example either in parentheses or in its own sentence).}
To do so, we calculate the empirical distribution for all $n$-grams %document frequency 
under style $p$. With this, we can calculate $f_{p,n}$'s probability in Equation \ref{eq:ngram2}, where $ECDF(f_{p,n})$ is the empirical distribution function which converts a document frequency $f_{p,n}$ to a probability under empirical distribution. 
we call this the \textit{ecdf score} for short in the discussion below. $F_{p,n}$ is a list that contains all $n$-gram document frequencies under style $p$.

To check how frequently $t$ occurs in documents of different styles, we calculate the number of stylized corpora $D_{s}$ where t has occurred (Equation \ref{eq:ngram3}). 
%in Equation \ref{eq:ngram3}, we calculate the number of stylized corpus $D_{s}$ where t has occurred. 
Here $S$ is the style set, which contains all styles. We use $occur\_num$ from Equation \ref{eq:ngram3} to penalize the score of t in Equation \ref{eq:ngram4}. If t has occurred in multiple corpora, then it is not deemed significant enough to represent the current style. 
%we consider it as not significant for representing current style.
To see how much better $t$ aligns with style $p$ than another style $q$, we finally calculate the contrastive $n$-gram score for $t$ under style $p$ by calculating the average difference between the ecdf score under style $p$ and all other styles $q$, where $q$ is in the style set $S$ but $q$ is not equal to $p$ (Equation \ref{eq:ngram4}). \\
\textbf{Bounds Analysis} The best case scenario of $CNG_{p,n}$ score for a $n$-gram $t$ is when $t$ only occurs in $D_{p}$ but not in any other $D_{q}$ and occurs in every sentence $d \in D_{p}$. That is to say $t$ is only associated with style $p$ and is also associated with every sentence in $D_{p}$. With this, we can calculate the upper bound for $CNG_{p,n}$, where
$CNG_{p,n} = \frac{\|S\|-1}{\|S\|} $. The worst case scenario is when $t$ never occurs in $D_{p}$ but occurs in all other datasets $D_{q},q!=p$ and appears in every sentence $d \in D_{q}$. This results in the lower bound for $CNG_{p,n}$, where $CNG_{p,n} = -\frac{1}{\|S\|} $. Notably, the stop words are naturally de-emphasized by $occur\_num$ and subtraction if they are nearly evenly distributed between different styles through Equation \ref{eq:ngram4}.
\subsection{OnlyStyle}
After we calculate the contrastive $n$-gram score for all $n$-grams in the dataset for a given style $p$, we can calculate how much a caption $c$ aligns with a style $p$. Since this score is designed to assess a single caption without any reference caption, we call it the \textit{OnlyStyle} score. Specifically, for any caption $c$, we first extract all the $n$-grams from it, where $n \in \left\{1,2,3,4 \right\}$. With these $n$-grams, we can compute the OnlyStyle score as follows:
\begin{align}
& OnlyStyle(c,p) = \frac{1}{4}\sum^{4}_{n=1} OnlyStyle_{n}(c,p) \label{eq:onlystyle1}\\
& OnlyStyle_{n} (c,p) = \frac{1}{\|M\|}\sum_{i \in M}CNG_{p,n}^{i} \label{eq:onlystyle2}
\end{align}
In Equations \ref{eq:onlystyle2}, we calculate the average CNG score across all $n$-grams from caption $c$, where $CNG_{p,n}^{i}$ is the CNG score for the $i$th $n$-gram under style $p$ and M is the set of all the $n$-grams in caption $c$. We get the OnlyStyle score using Equation \ref{eq:onlystyle1} for $c$ under style $p$ by averaging over $n$ of $OnlyStyle_{n}(c,p)$.\\
\textbf{Bounds Analysis} OnlyStyle displays the global view of a sentence's association with a style $p$ by averaging across different $CNG$ scores, so the bounds stay the same as for $CNG$. Specifically, suppose we could always get the highest $CNG_{p,n}$ for every $n$-gram in $M$, this will give us the highest OnlyStyle score for $c$ under $p$, which is $\frac{\|S\|-1}{\|S\|}$. On the contrary, if we always get the lowest $CNG_{p,n}$ for every $n$-gram in $M$, this will result in the lowest OnlyStyle score, which is $ -\frac{1}{\|S\|} $.
\subsection{StyleCIDEr}
While the ability to evaluate a caption without a reference sentence has its benefits, the presence of a reference sentence allows us to better evaluate the impact that individual words have. 
This is especially useful for stylized image captioning because individual words can have a large impact on the style that a sentence exhibits. 
To that end, we propose StyleCIDEr, a metric meant to evaluate how well a generated caption aligns with a reference caption, which emphasis placed on stylized words. 
This metric uses CIDEr as a kernel function. 
To better understand this, we will first outline how to calculate CIDEr. 
Suppose that we have two captions, $c_{i},c_{j}$:

\begin{align}
& CIDEr(c_{i},c_{j}) = \frac{1}{4}\sum^{4}_{n=1} CIDEr_{n}(c_{i},c_{j}) \label{eq:cider1}\\
& CIDEr_{n} (c_{i},c_{j}) = \frac{g^{n}(c_{i})\cdot g^{n}(c_{j})}{\|g^{n}(c_{i})\|\cdot \|g^{n}(c_{j})\|} \label{eq:cider2}
\end{align}
where $g^{n}(c_{i})$ is a vector where each element is formed by a $n$-gram score calculated by TF-IDF. TF-IDF is good at scoring salient $n$-grams for current captions, but it does not necessarily focus on stylized words or $n$-grams. As a result, we replace this TF-IDF score with our CNG score. This means that each element in $g^{n}(c_{i})$ will be a score focusing on a given style, like $p$. Then we calculate using Equations \ref{eq:cider1} and \ref{eq:cider2}, and we can compare $c_{i},c_{j}$ based on their stylized elements.\\
\textbf{Bounds Analysis} Since CIDEr score is bounded from (0,1), StyleCIDEr has the same range. In the best case scenario, when evaluating a caption that is the exact same as the reference caption, the StyleCIDEr would be 1.
If there are no overlapping $n$-grams between the generated caption and the reference caption caption, the StyleCIDEr score would be 0. To achieve a score of 1, StyleCIDEr does not require every $n$-gram in the generated caption to be the same as in the reference caption.
As long as $n$-grams related to current style overlap between the generated caption and the reference caption, the StyleCIDEr score can be 1. Also, if the generated caption and reference caption have several overlapping $n$-grams but none of them are related to current style, the StyleCIDEr can be 0.

\section{Experiment} 
To verify our proposed metrics, we evaluate them on two Datasets: PERSONALITY-CAPTIONS, which contains 215 different styles, and FlickrStyle10K, which contains 2 styles. 
In total, we perform three different evaluations. 
%To evaluate our metric is effective at scoring on the style parts for a caption, we perform three evaluations: 
First, we measured how well every ground truth caption performs compared to its ground truth style using both OnlyStyle and StyleCIDEr. 
We compared these values against a baseline in which ground truth captions were paired with a random style, rather than their respective ground truth style. 
The second evaluation involves re-evaluating popular frameworks for stylized image captioning using our proposed metrics. 
Compare how our metrics perform versus common NLP metrics such as BLEU and CIDEr. 
Finally, we perform a human subjects evaluation on OnlyStyle to verify that our metric aligns with human perceptions of style. 
\subsection{Datasets}
Since PERSONALITY-CAPTIONS and FlickrStyle10K are frequently used in stylized image captioning work \cite{shuster2019engaging,li20213m,gan2017stylenet,chen2018factual,guo2019mscap,zhao2020memcap}, we decide to validate our proposed metrics on these two datasets. 
The PERSONALITY-CAPTIONS dataset was released in 2019 \cite{shuster2019engaging}. 
It contains ground truth captions that were collected via human crowdsourcing. 
During data collection, humans are required to create engaging captions based on the image context and the given personalities so that the caption accurately embodies the given personality. 
%show their understanding of the personality.
Each data entry within this dataset is represented as a triple containing an image, personality trait, and caption. 
The images are selected from YFCC100M dataset~\cite{thomee2016yfcc100m}. 
In total, 241,858 captions are included in this dataset. 
Each caption is associated with one of the 215 personality traits selected from a list of 638 traits~\cite{gunkel2013638}. Even though the number of potential styles in the dataset makes captions more diverse, some of the styles in this dataset could be difficult for a human to distinguish (e.g. ``Happy" versus ``Cheerful").

In this work, we use the accessible data from this dataset which contains 186698 examples in the training set, 4993 examples in the validation set, and 9981 examples in the test set. 
The PERSONALITY-CAPTIONS dataset is large, which enables our contrastive $n$-gram metric to learn current style characteristics in a broader range. 
This naturally leads to the question of how our metric would perform with significantly fewer styles. To answer this question, we explore FlickerStyle10K.

FlickrStyle10K contains two styles: Humorous and Romantic. 
The ground truth stylized captions included in this dataset are based on factual image captions (descriptions without any style) that were modified so that they exhibit the given style. 
Thus, humorous captions and romantic captions are often very similar when describing the image, but different when using words that exhibit the desired style. 
%and different on some linguistic stylized words. 
Since only 7000 images are publicly available, we will use this FlickrStyle7K dataset in our experiment. 
When training a stylized captioning model, we use the protocol outlined in \cite{guo2019mscap,zhao2020memcap,li20213m} did. 
This involves first randomly select 6,000 images as the training data and using the remaining 1000 images as testing data. We further split 10\% data from the training set into a validation set. 

\subsection{Stylized Captioning} 
UPDOWN\cite{shuster2019engaging}, Multi-UPDOWN\cite{li20213m} and Visual Language Pretraining (VLP) on transformers\cite{zhang2021vinvl} have shown good performance captioning tasks (stylized or otherwise) when evaluated using common automated metrics (BLEU, CIDEr, etc.). 
As such, we evaluate these three types of models using our proposed metrics.

For the UPDOWN captioning model, we use the parameters outlined in \cite{shuster2019engaging}. 
We use pretrained ResNext spatial (7*7*2048) features and mean-pooled features (2048), along with style as one-hot vector as inputs to this network.
For the Multi-UPDOWN as captioning model, we use the process outlined in \cite{li20213m} and use ResNext spatial (7*7*2048) features and mean-pooled features (2048), 5 dense caption, along with style as a one-hot vector as inputs into its network. For SVinVL as a captioning model, we fine-tune VinVL on the PERSONALITY-CAPTION dataset by replacing object tags with 5 dense captions and styles (in text form) to learn the connection between image captions with ResNext spatial features (7*7*2048), dense captions, and styles. For each caption model, we also train a model with the same settings but without style inputs. So, in total, we train 6 models for each dataset and they are named:
(1) UPDOWN; (2) UPDOWN\_NoStyle; (3) Multi-UPDOWN; (4) Multi-UPDOWN\_NoStyle; (5) SVinVL; (6) SVinVL\_NoStyle.

For (1)-(4), we use entropy as loss function and Adam optimization with an initial learning rate of 5e-4. 
The learning rate decays every 5 epochs. 
For (5)-(6), we follow the fine-tuning steps of VinVL and use masked token loss and AdamW optimaztion with an initial learning rate of 3e-5. A linear scheduler is used for decaying the learning rate.
In total, we train 30 epochs when using the PERSONALITY-CAPTIONS dataset~\cite{shuster2019engaging} with a batch size of 128 and evaluate the model every 3000 iterations. We train for 100 epochs when using the FlickrStyle10K dataset \cite{gan2017stylenet} following the process outlined in \cite{shuster2019engaging,li20213m,li2020oscar}.
\begin{table*}
 \begin{center}
 \begin{tabular}{lll}
 \hline
   Dataset     & OnlyStyle Score & StyleCIDEr
    Score \\
    \hline
    PERSONALITY-CAPTIONS & 0.9775 &   0.9484 \\
    Flickr7k     & 0.9994 & 0.9032 \\ 
     \hline
  \end{tabular}
  \end{center}
  \caption{Evaluation of ground truth captions using OnlyStyle and StyleCIDEr following Equations \ref{eq:onlystyle_eval} and \ref{eq:stylecider_eval}}
  \label{tab:gdeval}
\end{table*}

\subsection{Evaluation}%%% below is table for flik
We will evaluate our proposed metric under three cases. First, we directly evaluate the ground truth captions in the dataset since each of them is associated with a style. %We consider this case as having ground truth. 
%brent{Not sure I understand this.}
Second, we evaluate different caption models and compare them with models that do not contain any style inputs. 
%without style inputs. 
Third, we perform a human study and ask them to evaluate some generated captions from a model. We then compare the human evaluation results with our proposed metric results. More details are included below.
\begin{table*}
  \begin{center}
    \begin{tabular}{lllllll}
  \hline
   Model & BlEU1 & BlEU4   & CIDEr     & StyleCIDEr & Onlystyle \\
   \hline
   UPDOWN & 0.241 & 0.034 & 0.231 &  \textbf{0.046} & \textbf{0.126} \\
   UPDOWN\_NoStyle & \textbf{0.255} & \textbf{0.035} &  \textbf{0.249} & 0.034 & 0.055 \\
  \hline
  \hline
 Multi-UPDOWN & 0.251 & 0.035 & \textbf{ 0.275} & \textbf{0.048}  & \textbf{0.094}   \\
 Multi-UPDOWN\_NoStyle & \textbf{0.252} & \textbf{0.037} & 0.261  & 0.026    & 0.049  \\
 \hline
 \hline
   SVinVL & 0.232 & 0.027    & 0.248     & \textbf{0.048} &  \textbf{0.129} \\
 SVinVL\_NoStyle & \textbf{0.233} & \textbf{0.029} &  \textbf{0.249} & 0.033 & 0.084  \\
  \hline
  \end{tabular}
  \end{center}
  \caption{Different Model Performance on
  \textbf{FlickrStyle7K}}
   \label{tab:flikres}
\end{table*}

\subsubsection{Evaluating Ground Truth Captions}
Each ground truth caption included in each dataset is associated with a given style. 
Because of this, we expect our metric could recognize if a caption was correctly paired with its ground truth style as opposed to a randomly selected style. 
%one caption is closer to the style it is associated than other styles. T
Thus, we would expect to see the following behaviors:
\begin{align}
&  \pmb{OnlytStyle}(s,p)>\pmb{OnlytStyle}(s,q) \label{eq:onlystyle_eval} \\
&  \pmb{StyleCIDEr}(s,s_{p})>\pmb{StyleCIDEr}(s,s_{q}) \label{eq:stylecider_eval} 
\end{align}
Here, $s$ is a caption from dataset and its gound truth style is $p$. $q$ is anther style and $q!=p$. 

With the above assumptions, we calculate the OnlyStyle score for each ground truth sentence for all styles in each dataset. We also calculate the StyleCIDEr score for each caption using either 1) all captions with the same style or 2) all captions with a different style as reference sentences. 
 
Then we calculate the number of sentences satisfying equation~\ref{eq:onlystyle_eval} and~\ref{eq:stylecider_eval} and average the results over all captions in each dataset. The results are reported in Table ~\ref{tab:gdeval}.

\subsubsection{Model Comparison}
If a trained stylized image captioning model were given only image features without any indication of what style should be generated, the model is likely to accurately describe the image, but have an unexpected style. 
In these situations, we would expect that our metrics to score higher on models that have this style knowledge provided in some way. 

To evaluate this, we use our metrics to evaluate the performance of the stylized image captioners described previously with their NoStyle equivalents. % for each caption model in our experiment, we compare their performance with the corresponding model without style inputs. For example, we compare UPDOWN model with UPDWON\_NoStyle model. 
For each model, we report their respective CIDEr, BLEU1, and BLEU4 to show the accuracy of the generated text. At the same time, we report OnlyStyle score and StyleCIDEr score for each model to show how well each generated caption aligns with its desired style. 
%much the model contributing on stylized generations on their stylized parts. 
The results for PESONALITY-CAPTIONS is in Table ~\ref{tab:personres} and the results for FlickrStyle7K is in Table ~\ref{tab:flikres}.

\subsection{Human Evaluation}
To evaluate the consistency between human judgments and our proposed metrics, we perform a human evaluation by asking humans to rank some sample generations from one of our models based on how well it represents a given style. 
For this evaluation, we focus solely on OnlyStyle since StyleCIDEr is based on the CIDEr metric, which has already been shown to align with human judgment. \\
\textbf{Human Study Setup}
For this experiment, we chose the following six styles from the PERSONALITY-CAPTIONS dataset: Abrasive (Annoying, Irritating), Angry, Curious, Fearful, Gloomy, Happy.
We chose these styles because they closely resembled the six basic emotions cited in Psychology literature~\cite{ekman1971constants}, and we felt that this would make it easier for users to identify them.

For each of these styles, we sample 3 captions generated by the Multi-UPDOWN model on test data. 
After approved by Institutional Review Boards,
we then asked crowdsourced users from Prolific to rank three generated sentences according to how well they are associated with a given style. 
Users provided answers in the range of 1-3 with 1 indicating the worst association, and 3 indicating the best association. 
Users were asked to perform this task 6 times, once for each style in the evaluation. 
In total, we received 96 responses to this task.

Note that this study and all the materials associated with it were reviewed and approved by our Institutional Review Board to ensure that subjects were exposed to no more than a minimal risk during their participation. 

\begin{table*}
  \begin{center}
  \begin{tabular}{lllllll}
  \hline
   Model & BlEU1 & BlEU4   & CIDEr     & StyleCIDEr & OnlyStyle \\
   \hline
   UPDOWN & \textbf{0.424} & \textbf{0.075} &\textbf{ 0.177} & \textbf{0.108} & \textbf{0.02}\\
  UPDOWN\_NoStyle & 0.287 & 0.024 & 0.082 & 0.029& 0.002 \\
\hline
  \hline
 Multi-UPDOWN & \textbf{0.430} & \textbf{0.081} & \textbf{0.183} & \textbf{0.116}  & \textbf{0.016}    \\
 Multi-UPDOWN\_NoStyle & 0.340 & 0.035 & 0.109  & 0.050    & 0.002  \\
  \hline
  \hline
   SVinVL & \textbf{0.416} & \textbf{0.063}    & \textbf{0.152}     & \textbf{0.115} &  \textbf{0.014} \\
 SVinVL\_NoStyle & 0.367 & 0.041 &  0.111 & 0.026 & 0.001  \\

  \hline
  \end{tabular}      
  \end{center}

  \caption{Model Performance on \textbf{PERSONALITY-CAPTIONS}}
    \label{tab:personres}
\end{table*}
\textbf{Consistency with Human Judgment}
\begin{table*}
  \begin{center}
  \begin{tabular}{lllllll}
  \hline
    CorrCoef\slash Style & Abrasive (Annoying, Irritating) & Happy & Gloomy & Curious & Angry &  Fearful\\
    \hline
   overall Pearson $\rho$ & 0.874 & 0.017 & 0.878 & 0.053 & 0.875 & 0.039\\
   overall Spearman  $\rho$ & 1.0  & 0.5 & 1.0 & 0.5 & 1.0 & 0.5\\
  \hline
  \end{tabular}      
  \end{center}
 \caption{Pearson and Spearman Correlation Between Human Judgment and OnlyStyle}
  \label{tab:pearson}
\end{table*}
We pick the most common ranking of the three sentences for each style from the 96 users as human's judgment. 
We then compare these rank scores with the OnlyStyle scores on the sample generations.
We use Pearson and Spearman correlation coefficients to quantify the consistency between our proposed OnlyStyle metric and human judgments. We report both scores for the six styles we examined in Table ~\ref{tab:pearson}.

\section{Results and Discussion}

\subsection{Evaluating Ground Truth Captions}
As we see in Table \ref{tab:gdeval}, all of the scores are above 90\%.
This means in most cases, Equation \ref{eq:onlystyle_eval} and Equation \ref{eq:stylecider_eval} are satisfied. 
This matches our expectations, which means that OnlyStyle can differentiate correct and incorrect styles, and StyleCIDEr can measure the accuracy of stylized word generation. However, we should point out that all ratings did not achieve 100\% performance. 
%notice all the ratings didn't achieve 100\%. 
This is because some sentences may not be representative of the style that they are associated with in the dataset. 
%might not that representative for the style it is labeled. 
As an example, it is possible that some sentences appear in ``Happy" and also appear in ``Cheerful". Despite this, some of these sentences might have a higher score in ``Cheerful" than ``Happy". So, when we evaluate $OnlyStyle(s,``Happy")> OnlyStyle(s,``Cheerful")$, Equation \ref{eq:onlystyle_eval} is not satisfied. In addition, when we evaluate $StyleCIDEr(s,s_{Happy})>StyleCIDEr(s,s_{Cheerful)})$, it is not satisfied either. 

With StyleCIDEr, it is also possible that the two captions used for evaluation do not have overlapping words, regardless of if they are drawn from the same style or different styles. 
In this situation, both the within-style StyleCIDEr score and between-style StyleCIDEr score would be zero, which violates Equation \ref{eq:stylecider_eval}. 

\subsection{Model comparison}
\textbf{With Style Versus Without.} From Tables \ref{tab:flikres} and \ref{tab:personres}, we can see that models with style modality as inputs have higher OnlyStyle and StyleCIDEr scores across all captioning models. This provides support to our claim that both StyleCider and OnlyStyle can measure whether a model can effectively generate output that aligns with a given style.\\
\textbf{Style Metric vs Accuracy Metric.} From Table \ref{tab:flikres}, we can clearly see that when evaluated using only BLEU and CIDEr, UPDOWN\_NoStyle and SVinVL\_NoStyle perform better than UPDOWN and SVinVL. One reason for this may be because the stylized captions in FlickrStyle7K are created by modifying the factual captions. In this situation, models without a style vector as an input would generate the same output for both romantic and humorous styles, which strongly resembles the original factual caption. 
These outputs would score highly when evaluated using CIDEr and BLEU. 

Models with style included as an input will produce different sentences to better exemplify the desired style. 
As a result, these generated captions may have fewer overlapping words than the reference sentence as they potentially transfer stylized words from other examples, which in turn results in lower BLEU and CIDEr scores. 
This shows that we can not simply rely on these metrics in this situation. 
\\
\textbf{Caption Model performance}
If we only look at models with style inputs, from Table \ref{tab:flikres}, we can see that the UPDOWN Model achieves the highest OnlyStyle score for PERSONALITY-CAPTIONS and ranks second on FlickrStyle7K, which means that the UPDOWN Model are good at generating more stylized sentences. 
However, the UPDOWN model has the worst score on StyleCIDEr. 
This means that while the UPDOWN model can produce stylized words that align with the desired style, they may not accurately describe the image in question.

Multi-UPDOWN has the best StyleCIDEr score as well as excellent performance on BLEU and CIDEr scores. 
This is likely because it is directly trained address limitations with UPDOWN where it would overly focus on building connections between styles and generated captions and miss information in the visual space. 
The Multi-UPDOWN model ranks second on PERSONALITY-CAPTIONS and is the worst performing model on FlickrStyle7K on OnlyStyle, which means that the gain in visual understanding likely came at the expense of generating stylized output.

The SVinVL model has the worst accuracy performance on both datasets. Its StyleCIDEr rating ranks as the best on FlickrStyle7K and the second on PERSONALITY-CAPTION. Its OnlyStle rating ranks the best on FlickrStyle7K and worst on PERSONALITY-CAPTION. This means that the SVinVL Model tends to connect style with caption generations on FlickrStyle7K which has 2 styles, and  struggles to make correct connections on the PERSONALITY-CAPTIONS dataset where size scales up. %We speculate  
\begin{figure*}
\centering
 \includegraphics[width=0.9\textwidth,height=3.3cm]{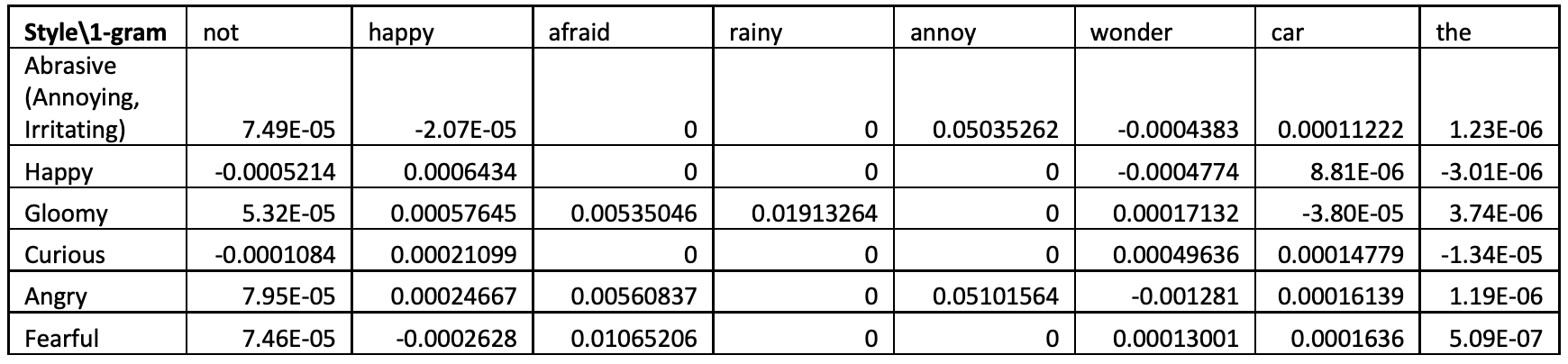}
\caption{The Contrastive 1-gram score for sample words under 6 styles; First column contains the sampled 6 styles; The 2-9 columns are the Contrastive 1-gram scores for the corresponding sampled word.}
\label{fig:notprob}
\end{figure*}
\subsection{Human Evaluation}
In Table \ref{tab:pearson}, we can see Pearson and Spearman correlation coefficients are very high with style Abrasive (Annoying, Irritating), Gloomy and Angry, which indicates that rankings determined using the OnlyStyle metric are consistent with rankings derived based on human judgment.  
%consistency between human judgment and OnlyStyle scores is high. 
However, Pearson and Spearman scores are low for Happy, Curious and Fearful. 
Because of this finding, we decided to further investigate this discrepancy. 

%With this, we further investigate the disagreement of these styles.
Recall that for each style, we asked users to rank three generated sentences.
By investigating how users ranked these sentences, we hope to uncover the source of any ranking discrepancy. 
We found that for the ``Happy" style, people ranked the sentence ``what a beautiful day" higher than ``i love rugby", which means they think the first sentence is more associated with ``Happy" than the latter one. OnlyStyle scores these sentences in the opposite way. After further investigation, we discovered that the phrase, ``what a beautiful day," appears in multiple styles in the dataset (e.g. ``Warm", ``Cheerful", and ``Enthusiastic") but only occurs in ``Happy" as a part of a larger sentence. The phrase, ``i love rugby," only occurs in ``Happy" as a complete sentence. 

For ``Curious", people rank ``i wonder what these people are thinking about" higher than ``i wonder how many lamps there are" and OnlyStyle scores the opposite. 
The first sentence never appears in ``Curious" and only appears in ``Freethinking" in the dataset. The second sentence only appears in ``Curious". 

For ``Fearful", people rank the caption, ``i hope he doesn t fall," higher than the caption, ``i hope that statue doesn t fall". The first sentences appears in ``Fearful" and also appears in ``Sympathetic" and ``Gloomy" in the dataset. The second sentence only appears in ``Fearful". 

Even though people's judgments are very reasonable, OnlyStyle gives higher scores to phrases that appear frequently in one specific style, with the assumption that those are more indicative of a given style.  
To investigate a little further, we calculate the OnlyStyle scores for these sentences on all 215 styles in the PERSONALITY-CAPTIONS dataset. We then calculate the retrieval ranks for ``Happy", ``Curious", and ``Fearful" for their respective sentences. We found that for each of the disagreeing sentences, the correct style was always in the top 10\% of retrieved styles, indicating that these styles can be difficult to distinguish from other top styles.
It also means that even though these sentences are ranked differently by humans and the OnlyStyle metric, they are all very close the given styles. Thus, it would be hard for human to differentiate them and rank them as well. Overall, we think all these disagreements could be resolved with better datasets.

\subsection{Additional Discussion}
We mentioned previously that basing an evaluation metric on TF-IDF can be problematic. 
Some words, such as ``not'' or ``happy'' can appear across all the styles, which would be scored as 0.
We hypothesized that our use of a contrastive $n$-gram score could address some of these issues. 
We evaluated this by sampling several words including``not'' and ``happy'' under six styles and calculating their contrastive 1-gram score (shown in Figure~\ref{fig:notprob}). 

We can see that ``not" generally scores very low due to its occurrence in every style in the dataset; however, it scores negatively in "Happy" and "Curious", which we view as a positive result.
The 1-gram, "happy," scores negatively in ``Abrasive (Annoying, Irritating)" and ``Fearful" and scores highest in the style, "Happy". Figure~\ref{fig:notprob} also shows CNG deprecates word like ``the" and doesn't neglect noun's contribution to style, like ``car".  These all shows that using the contrastive $n$-gram score allows us to get a more nuanced understanding of how $n$-grams contribute to different styles.

\section{Conclusion}
In this paper, we propose two metrics for evaluating stylized image captioning models: OnlyStyle and StyleCIDEr.
These metrics can be used to automatically measure the association between a caption and a given style. We have shown through a series of evaluations that these metrics are effective at evaluating stylized image captions and align with how humans perceive stylized sentences. 

Despite the effectiveness of these metrics, they are still based on evaluating $n$-grams. 
As such, they are subject to the same limitations of similar methods (e.g. BLEU and CIDEr). The primary limitation of these types of models is that $n$-grams might fail to capture distant dependencies. \cite{isozaki2010automatic}.

There are also potential ethical implications one must consider with these approaches as well. 
By understanding how these metrics work, it is possible that an adversary could use this knowledge to better their generation of malicious captions that still perform well on these metrics.
This may enable them to generate undesirable content that still scores well in practice. 

Despite these limitations, we feel that these metrics will be useful to researchers in helping them evaluate stylized image captioning models where traditional NLP metrics may not perform well.

%%%%%%%%% REFERENCES
{\small
\bibliographystyle{ieee_fullname}
\bibliography{egbib}
}

\end{document}